\documentclass[letterpaper, 10 pt, conference]{ieeeconf} 
\IEEEoverridecommandlockouts                              

\usepackage{amsmath}
\usepackage{amssymb}  
\usepackage{amsfonts}
\usepackage{mathtools}
\usepackage{bm}
\usepackage{color}
\usepackage{graphics} 
\usepackage{graphicx}
\usepackage{epsfig} 
\usepackage{epstopdf}
\usepackage[noadjust]{cite}
\usepackage{tikz}
\usetikzlibrary{calc,positioning,shapes,shadows,arrows,fit}
\usepackage{float}
\usepackage[font = footnotesize]{caption}
\usepackage{algorithm}
\usepackage{algorithmic}
\usepackage{balance}

\usepackage{amsthm}

\theoremstyle{definition}
\newtheorem{definition}{Definition}
\newtheorem{remark}{Remark}

\renewcommand{\baselinestretch}{0.97}

\title{\LARGE \bf Safe RAN control: A Symbolic Reinforcement Learning Approach}

\author{Alexandros Nikou$^{1}$, Anusha Mujumdar$^{1}$, Vaishnavi Sundararajan$^{2}$, Marin Orlic$^{1}$ and Aneta Vulgarakis Feljan$^{1}$
\thanks{$^{1}$The authors are with Ericsson Research, Research Area Artificial Intelligence (AI). Link: https://www.ericsson.com/en/future-technologies. Contact authors' e-mails: {\tt\small \{alexandros.nikou, anusha.pradeep.p.mujumdar, marin.orlic, aneta.vulgarakis\}@ericsson.com}}%
\thanks{$^{2}$The author is with University of California Santa Cruz. E-mail: {\tt\small vasundar@ucsc.edu}}%
}

\begin{document}

\maketitle
\thispagestyle{empty}
\pagestyle{empty}

\begin{abstract}
In this paper, we present a Symbolic Reinforcement Learning (SRL) based architecture for safety control of Radio Access Network (RAN) applications. In particular, we provide a purely automated procedure in which a user can specify high-level logical safety specifications for a given cellular network topology in order for the latter to execute optimal safe performance which is measured through certain Key Performance Indicators (KPIs). The network consists of a set of fixed Base Stations (BS) which are equipped with antennas, which one can control by adjusting their vertical tilt angles. The aforementioned process is called Remote Electrical Tilt (RET) optimization. Recent research has focused on performing this RET optimization by employing Reinforcement Learning (RL) strategies due to the fact that they have self-learning capabilities to adapt in uncertain environments. The term safety refers to particular constraints bounds of the network KPIs in order to guarantee that when the algorithms are deployed in a live network, the performance is maintained. In our proposed architecture the safety is ensured through model-checking techniques over combined discrete system models (automata) that are abstracted through the learning process. We introduce a user interface (UI) developed to help a user set intent specifications to the system, and inspect the difference in agent proposed actions, and those that are allowed and blocked according to the safety specification. 

\end{abstract}

\section{Introduction}

There exists a push for future generations of mobile networks, such as 6G, to leverage AI in their operations. Simultaneously, future cellular networks are expected to be exceedingly complex, and demand real-time and dynamic network optimization and control. This constitutes one of the key challenges for network operators. It is desirable that network configuration is optimized automatically and dynamically in order to:
\begin{itemize}
\item satisfy consumer demand, with User Equipment (UEs) highly distributed in both spatial and temporal domains,
\item account for the complex interactions between multiple cells that shape the KPIs of a region in the network, and finally,
\item  ensure acceptable Quality of Experience (QoE) to each UE in the network
\end{itemize}
In such a scenario, the objective is to optimize a set of network KPIs such as \emph{coverage}, \emph{quality} and \emph{capacity} and to guarantee that certain bounds of these KPIs are not violated (safety specifications). The optimization can be performed by adjusting the vertical electrical tilt of each of the antennas of the given network, known in the literature as the RET optimization problem (see \cite{guo2013spectral, razavi2010fuzzy, fan2014self, buenestado2016self, balevi2019online, yilmaz2010self}). For example, an increase in antenna downtilt correlates with a stronger signal in a more concentrated area as well as higher capacity and reduced interference radiation towards other cells in  the  network.  However,  excessive  downtilting  could also  lead to insufficient coverage in a given area, with some UEs unable to receive a minimum Reference Signal Received Power (RSRP). Existing  solutions  to  downtilt  adjustment in the industry use  rule-based  algorithms  to  optimise  the  tilt  angle  based  on   historical  network  performance.  These  rules  are  usually  created by domain experts, and thus lack the scalability and adaptability  required  for  modern  cellular  networks.  

Reinforcement learning (RL) \cite{sutton2018reinforcement, mnih2015human, bouton2020point, zavlanosRL} has become a powerful solution for dealing with the problem of optimal decision making for agents interacting with uncertain environments. It is widely known that RL performs well on deriving optimal policies for optimizing a given criterion encoded via a reward function, and can be applied in many use cases such as robotics, autonomous driving, network optimization, etc. \cite{polydoros2017survey, luong2019applications}. However, it is also known that the large-scale exploration performed by RL algorithms can sometimes take the system to unsafe states \cite{garcia2015comprehensive}.

Considering the problem of RET optimization, RL has been proven to be an efficient framework of KPI optimization due to its self-learning capabilities and adaptivity to potential environment changes \cite{vannella2020off, vannella2020remote}. For addressing the safety problem (i.e., to guarantee that the desired KPIs remain in certain specified bounds) authors in \cite{vannella2020off, vannella2020remote, feghhi2020safe} have proposed a statistical approach to empirically evaluate the RET optimization in different baseline policies and in different worst-case scenarios. With this approach, safety is defined with respect to a minimum performance level compared to one or more safety baselines that must be ensured at any time.

The aforementioned statistical approach focuses on ensuring the reward value remains above a desired baseline and do not provide a mechanism of blocking actions that violate undesired system behavior. In particular, a more powerful notion of safety can be expressed in terms of safe states or regions, defined according to a (formal) intent specification \cite{fulton2018safe}. Such an approach decouples the notion of safety from that of reward. Intuitively, safety intents define the boundaries within which the RL agent may be free to explore. Motivated by the abovementioned, in this work, we propose a novel approach for guaranteeing safety in the RET optimization problem by using model-checking techniques and in parallel, we seek to generalize the problem in order to facilitate richer specifications than safety. In order to express desired specifications to the network into consideration, Linear Temporal Logic (LTL) (see \cite{katoen, loizou_2004, alex_automatica_2017, alex_PhD}) is used, due to the fact that it provides a rich mathematical formalism for such purpose. Our proposed framework exhibits the following attributes:
\begin{itemize}
\item a general automatic framework from LTL specification user input to the derivation of the policy that fulfills it; at the same time,  blocking the control actions that violate the specification; 
\item novel system dynamics abstraction to Markov Decision Process (MDP) which is computational efficient;
\item UI development that allows a user to graphically access, understand and trust the steps of the proposed approach.
\end{itemize}

\begin{table}[t!]
\begin{center}
\begin{tabular}{|c||c|}
\hline
ANN & Artificial Neural Network \\
\hline
BA  & B\"uchi Automaton \\		
\hline
CMDP & Companion Markov Decision Process \\
\hline
DQN & Deep Q - Network \\
\hline
KPIs & Key Performance Indicator \\
\hline
LTL  & Linear Temporal Logic \\
\hline
MDP  & Markov Decision Process \\
\hline
QoS & Quality of Service \\
\hline
RET & Remote Electrical Tilt \\		
\hline
RBS  & Radio Base Station \\		
\hline
RAN  & Radio Access Network \\		
\hline
RRC & Radio Recourse Control \\
\hline
RSRP & Reference Signal Received Power \\
\hline
SGD & Stochastic Gradient Descent \\
\hline
TA & Timing Advance \\
\hline
SRL  & Symbolic Reinforcement Learning \\
\hline 
UI & User Interface \\
\hline
UE & User Equipment \\
\hline 
\end{tabular}
\end{center}
\caption{List of acronyms}
\label{table_parameters2}
\end{table}

\emph{\textbf{Related work}.} A framework that handles high-level specifications to RL agents is proposed in \cite{icarte2018using}. However, such approach requires the development of reward machines, which requires significant engineering that requires effort and knowledge, and it cannot be handled in an automated way in the sense that if the environment or the use case changes, new reward function development is required. Authors in \cite{alshiekh2018safe} propose a safe RL approach through shielding. However, the authors assume that the system dynamics abstraction into an MDP is given, which in the network applications that this manuscript refers to is challenging. For a comprehensive survey of safe RL we refer to \cite{garcia2015comprehensive}. As mentioned previously, authors in \cite{vannella2020off} address the safe RET optimization problem, but rely on statistical approaches that cannot handle general LTL specifications that we treat with this manuscript. A preliminary short version of this paper without deep technical details is shown in a demo track in \cite{nikou2021symbolic}.

This manuscript is structured as follows. Section \ref{sec:notation_preliminaries} gives some notation and background material. In Section \ref{sec:main_results}, a detailed description of the proposed solution is given. Section \ref{sec:demonstration} develops a concrete solution using the proposed approach in the RET optimization problem, through a video and UI, and experimental results. Finally, Section \ref{sec:conclusions} is devoted to conclusions and future research directions.

\section{Notation and Background} \label{sec:notation_preliminaries}

Throughout this manuscript, the abbreviations listed in Table \ref{table_parameters2} will be used. In the sequel, we review some background material from model checking theory and RL based RAN control.

\subsection{Linear Temporal Logic (LTL)}

In this paper we focus on task specifications $\varphi$ given in LTL. The syntax of LTL (see \cite{katoen}) over a set of atomic propositions $\Pi$ is defined by the grammar:
\begin{equation*}
\varphi := \top \ | \ \varpi \ | \ \neg \varphi \ | \ \varphi_1 \wedge \varphi_2 \ | \ \bigcirc \varphi  \ | \  \varphi_1 \ \mathcal{U} \ \varphi_2,
\end{equation*}
where $\varpi \in \Pi$ and $\bigcirc$, $\mathcal{U}$ stand for the next and until operators, respectively; $\neg$ and $\wedge$ are the negation and conjunction operator respectively. The always ($\square$) and eventually ($\Diamond$) operators can be defined by $\square  \coloneqq \neg \Diamond \neg \varphi, \ \ \Diamond \coloneqq \top \mathcal{U} \varphi$, respectively. LTL can be used to express any type of temporal tasks for dynamical systems \cite{alex_PhD}.

\begin{definition} \label{def: NBA}
A \textit{B\"uchi Automaton (BA)} is a tuple $(Q, Q_0, 2^\Pi, \delta, F)$ where 
\begin{itemize}
\item $Q$ is a finite set of states;
\item $Q_0 \subseteq Q$ is a set of initial states;
\item $2^\Pi$ is the alphabet;
\item $\delta : Q \times 2^\Pi \to 2^Q$ is a transition relation;
\item $F \subseteq Q$ is a set of accepting states.	
\end{itemize}
\end{definition}

It has been proven that every LTL formula can be translated to a BA that models all the system runs satisfying the formula (see \cite{gastin2001fast} for fast LTL to BA translation tools).

\begin{figure}[t!]
\centering
\includegraphics[scale = 0.50]{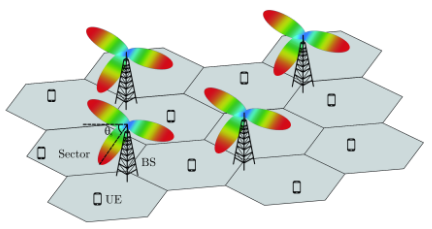}
\caption{A multi-cell wireless mobile network environment \cite{vannella2020off}.}\label{fig:env1}
\end{figure}

\begin{figure}[t!]
	\centering
	\includegraphics[scale = 0.40]{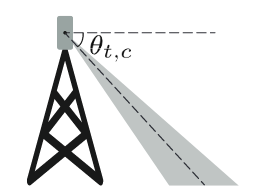}
	\caption{Representation of the downtilt $\theta_{t,c}$ at time $t$ for cell $c$.}\label{fig:antenna}
\end{figure}

\begin{figure*}
\centering
\includegraphics[height=8.0cm]{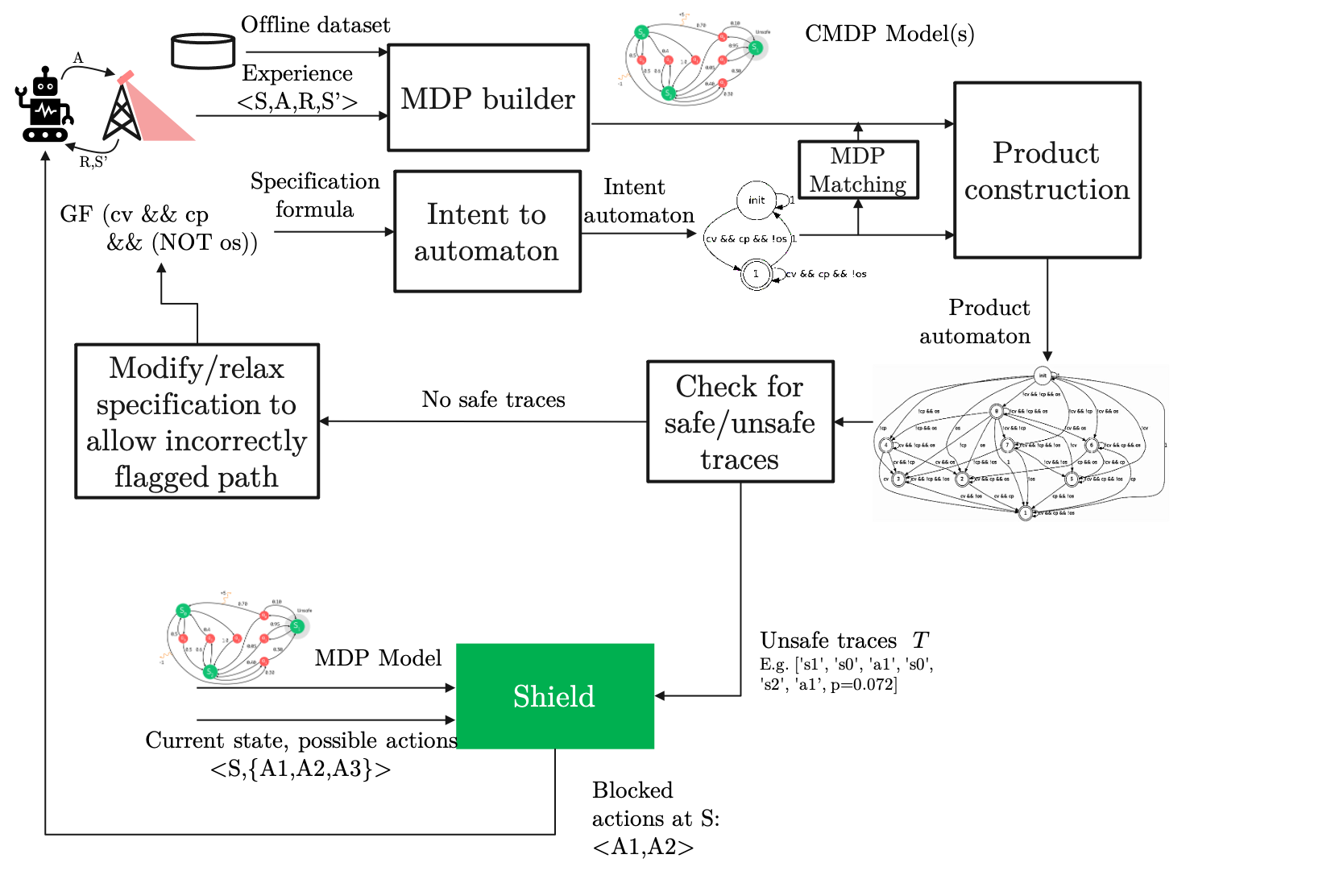}
\vspace{-0.1cm}
\caption{A graphical illustration of the sequence of steps of the proposed architecture.}
\vspace{-0.2cm}
\label{fig:test1}
\end{figure*}
\vspace{-0.2cm}

\subsection{Reinforcement Learning based RAN control}

Consider an area covered by $\mathcal{R}$ Radio Base Stations (RBS) with $\mathcal{C}$ cells that serve a set of $\mathcal{U}$ UEs uniformly distributed in the area (see Fig. \ref{fig:env1}). Denote by $\theta_{t, c}$ the antenna tilt of the cell $c \in \mathcal{C}$ at time $t \ge 0$, as depicted in Fig. \ref{fig:antenna}. The RET optimization problem has a goal to maximize network \emph{capacity} and \emph{coverage} while minimizing inter-cell interference. Such interference is modeled by the \emph{quality} KPI. The RET control strategy handles the antenna tilt of each of the cells, and is executed independently for each cell. In particular,
\begin{itemize}
\item the coverage KPI measures the degree to which a region of interest is adequately covered by the signal, and is computed using measurements of the Reference Signal Received Power (RSRP). 
\item the capacity KPI is an indicator of degree of congestion in the cell. This is calculated from the Radio Recource Control (RRC) congestion rate. 
\item the effect of \emph{negative cell interference} from neighboring cells is modeled by the quality KPI. The quality is calculated using the cell overshooting and cell overlapping indicators, which in turn depend on measurements of the RSRP level differences between a cell and its neighbors.
\end{itemize}

The RL agent observes a state of the environment, applies an action, receives a reward, and transitions to the next state \cite{sutton2018reinforcement}. The goal is to learn a policy that maximizes the cumulative reward over a time horizon. The environment of the RL agent is a simulated mobile network, and the system model is captured via a Markov Decision Process (MDP) $(\mathcal{S}, \mathcal{A}, \mathcal{P}, \mathfrak{R}, \gamma)$ that consists of:
\begin{itemize}
\item $\mathcal{S} \subseteq [0,1]^m$ discrete states that consist of normalized values for downtilt, and KPIs such as the (RRC) congestion rate, Timing Advance (TA) overshooting, coverage, capacity and quality, i.e., $s_{t,c} = [\theta_{t, c}, KPI_{t, c}] \in \mathcal{S}$, with state vector dimension $m$, where $KPI_{t, c}$ is a vector of state observations returned by the environment (in our use case the dimension of $KPI_{t, c}$ is between 15 and 45, depending on the features from the simulator selected to be part of the state). For example, 
\vspace{-0.1cm}
\begin{align}
  \phantom{i + j + k}
  &\begin{aligned}
    \mathllap{KPI_{t, c}} &=[{\rm COV}_{t,c}, {\rm CAP}_{t,c}, {\rm QUAL}_{t,c},...,KPI^k_{t, c}, ...\\
      & {\rm SINR}_{t,c}, {\rm TA}\_{\rm OS}_{t,c},
      {\rm RRC}\_{\rm CONG}\_{\rm RATE}_{t,c}]. \nonumber
  \end{aligned}
\end{align}

\item discrete actions $\mathcal{A} = \{-\alpha, 0, \alpha\}$ where $\alpha$ denotes the magnitude of the downtilt; At cell $c$ and at time $t$ the agent selects $a_{t,c} \in \mathcal{A}$; 
\item transition probability matrix $\mathcal{P}$ which describes the state evolution given the current and the executed by the action state;
\item scalar rewards $\mathfrak{R}$ that are the log squared sum of the coverage, capacity and quality, i.e.:
\begin{align*}
r_{t,c} = - \log\left( 1+{\rm COV}_{t,c}^2+{\rm CAP}_{t,c}^2+{\rm QUAL}_{t,c}^2 \right) \in \mathcal{R}.
\end{align*}
\item discount factor $\gamma \in [0, 1]$.
\end{itemize}

The policy of the RL agent $\pi: \mathcal{S} \to \mathcal{A}$ is a function that maps the states to actions that define the agent's strategy. At each discrete time instant $t$ the RL agent receives a state of the environment, selects an action, receives a reward and transits to a new state. The goal is to maximize the cumulative reward over a period of time.

\subsection{Q learning and Deep Q Network}

Q learning \cite{sutton2018reinforcement} is an RL learning algorithm that aims at estimating the state-value function:
\begin{align*}
Q^\pi(s,a) \coloneqq \mathbb{E} \left[ \sum_{t = 0}^{\infty} \gamma^t r_{t+1} | s_t = s, a_t = a\right],
\end{align*}
under a policy $\pi$, where $\mathbb{E}$ stands for the expected value. When an Artificial Neural Network (ANN) parameterization is chosen to estimate the Q function, we refer to the procedure as a Deep Q - Network (DQN); DQN uses experience replay memory $\mathcal{D} = \{(s_i, a_i, r_i, s'_i)\}_{i = 1}^{N}$, which is a means to store experience trajectories in datasets and use them for training purposes; the trajectories are enumerated for $i \in \{1, \dots, N\}$, where $N$ is the number of samples; DQN uses Stochastic Gradient Descent (SGD) methods with update $w \leftarrow w - \eta \cdot \nabla \left[y_t-Q(s_t, a_t) \right]^2$, where $\eta$, $y_t$, $\nabla$ stands for the learning rate, the target function at step $t$, and the gradient operator, respectively, in order to minimize the error between the target and parameterized Q function.

\begin{algorithm}[t!]
\caption{}
\begin{algorithmic}[1]
\STATE \textbf{Input:} LTL formula $\varphi$ given by the user
\STATE 
\STATE \textbf{Step 1}: Gather the experience replay data $\mathcal{D} = \{(s_i, a_i, r_i, s'_i)\}_{i=1}^{N}$ from simulation. Select subset of states as per intent;
\STATE \textbf{Step 2}:  Discretize selected states into $N_{b}$ bins, creating discretized states $\underline{\mathcal{S}}$. The size of state space is $|\mathcal{S}|^{N_b}$;
\STATE \textbf{Step 3}: Construct the CMDP $(\underline{\mathcal{S}}, \mathcal{A}, \mathcal{P}, \mathfrak{R}, \gamma)$;
\STATE \textbf{Step 4}:  Pass the LTL specification $\varphi$ to model checking \textbf{Algorithm 2};
\STATE \textbf{Step 5}:  Model checking returns traces that violate $\varphi$;
\STATE \textbf{Step 6}:  At each step in RL agent Function \textit{Shield}(MDP, T) blocks unsafe actions.
\end{algorithmic} 
\end{algorithm}

\section{Proposed Solution} \label{sec:main_results}

Our solution relies on a sequence of steps taken in order to match the LTL specification with the RL agent as it is depicted in Fig. \ref{fig:test1}, and block the actions that could lead the RL agent to unsafe states. 

Initially, the provided specification is converted to a BA as explained in Section \ref{sec:notation_preliminaries}. Then, by gathering experience data tuples from the RL agent, which is trained within a simulation environment with state-of-the-art model-free RL algorithms (DQN, Q-learning, SARSA \cite{guo2013spectral, razavi2010fuzzy, fan2014self, buenestado2016self}) we construct the system dynamics modelled as an MDP. In this solution, we develop a novel structure known as Companion MDPs (CMDPs); CMDPs do not encode the state transitions in terms of the full set of state features. E.g. consider that a state vector has features $s=[x_1, x_2, \dots, x_n]$. We select only a subset of features, e.g., $s_{cmdp, 1}=[ x_2, x_4, x_5]$. Similarly the next state vector would be $s'_{cmdp, 1} = [x'_2, x'_4, x'_5]$.  There could be several other companion MDPs with various subsets of state features e.g. $s_{cmdp, 2} =[ x_1,x_3,x_5 ]$, $s_{cmdp, 3} =[ x_3, x_5, x_6 ]$  etc. However, only the abstractions relevant to the specification is chosen, by matching the features in the specification to those in the companion MDP. Such an approach reduces the state space complexity, and retains only the relevant features depending on the intent. An MDP Matching component matches the intent to the relevant CMDP (depending on the features mentioned in the intent). In future work we would like to assess any loss of accuracy in predicting safety due to such an approach.\\

The experience data tuples generated over training are in the form $(s, a, r, s')$ where $s$ indicates the current state, $a$ indicates the executed action, $r$ the received reward that the agent receives after applying action $a$ at state $s$; and $s'$ stands for the state the agent is transitioned to after executing action $a$ at state $s$. In order to match the BA from the given LTL specification and the MDP, the states of the MDP are labelled according to the atomic propositions $\Pi$ from the LTL specification through a labeling function $L: \mathcal{S} \to 2^{\Pi}$. The atomic propositions set $\Pi$ consists of combination of KPIs in terms of low and high values. Then, by computing the product of the MDP with the specification, we construct an automaton $\mathcal{T} = {\rm MDP} \otimes \mathcal{A}_{\varphi}$ that models all the possible behaviours of the system over the given specification. At the same time, by negating the given formula, the automaton $\overline{\mathcal{T}} = {\rm MDP} \otimes \mathcal{A}_{\neg \varphi}$ is computed in order to compute any possible unsafe traces, which is done by applying graph techniques (such as Depth First Search algorithm see \cite{dfs}) on this automaton. In this way, we are able to compute the system traces that satisfy the intent; using these the actions that lead to violation of the intent can be blocked.
The abovementioned process is depicted more formally in Algorithm 1 and Algorithm 2. 

\begin{algorithm}[t!]
\caption{Model checking}
\begin{algorithmic}[1]
\STATE \textbf{Input:} LTL formula $\varphi$
\STATE \textbf{Output:} Unsafe traces that violate $\varphi$
\STATE 
\STATE \textbf{Step 1}: Translate the LTL formula to a BA $\mathcal{A}_{\varphi}$; Compute also the automaton $\mathcal{A}_{\neg \varphi}$.
\STATE \textbf{Step 2}:  Compute the product automaton $\mathcal{T} = {\rm MDP} \otimes \mathcal{A}_{\varphi}$ which essentially encodes all the possible behavior of the system;
\STATE \textbf{Step 3}:  Apply graph techniques to the product $\overline{\mathcal{T}} = {\rm MDP} \otimes \mathcal{A}_{\neg \varphi}$ for the calculation of any possible unsafe traces.
\end{algorithmic} 
\end{algorithm}

\begin{figure}[t!]
\vspace{-2mm}
\centering
\vspace{-2mm}
\includegraphics[scale = 0.40]{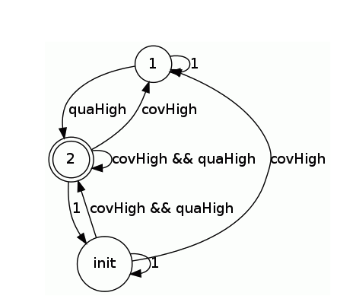}
\caption{The BA for the formula $\varphi_3 = \square(\Diamond \text{covHigh}) \wedge (\Diamond \text{qualHigh})$ with atomic propositions $\Pi = \{{\rm covHigh}, {\rm qualHigh}\}$. The state $2$ should be visited with actions $\rm covHigh$ and $\rm qualHigh$ over all futures.}\label{fig:ltl}
\end{figure}

\begin{figure}[t!]
\vspace{2mm}
\centering
\includegraphics[scale = 0.25]{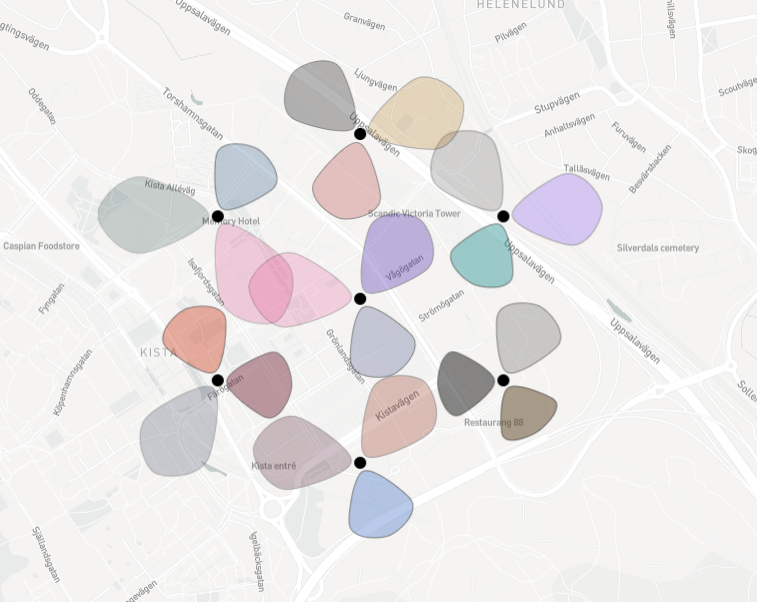}
\caption{A simulated mobile network with 21 cells.}
\label{fig:env2}
\end{figure}

Due to expressiveness of LTL, it may be possible that the user can choose an input intent that results in no safe traces for a given system. In such situations, the proposed technique cannot arrive at any solution, since any proposed action to the RL agent would lead to the violation of the given intent. For avoiding such configuration, i.e., in cases that there the executed algorithms results to no safe trace, we modify the given input to a new intent that results to some safe traces (see Fig. \ref{fig:env2}, inner loop). This procedure is currently performed by trial and error, and we are currently investigating how this re-configuration process can be performed in an automatic way.

\begin{remark}
It should be mentioned that our proposed architecture is general, and can be applied to any framework in which the dynamical system under consideration is abstracted into an MDP (see Section \ref{sec:notation_preliminaries}), for which LTL specifications need to be fulfilled. For example, in robot planning applications, the states are locations of the environment that the robot can move, and atomic propositions are the goal state and the obstacles. The potential LTL formula in such a scenario would include reachability and safety tasks.
\end{remark}

\begin{figure}[t!]
\vspace{2mm}
\centering
\includegraphics[scale = 0.5]{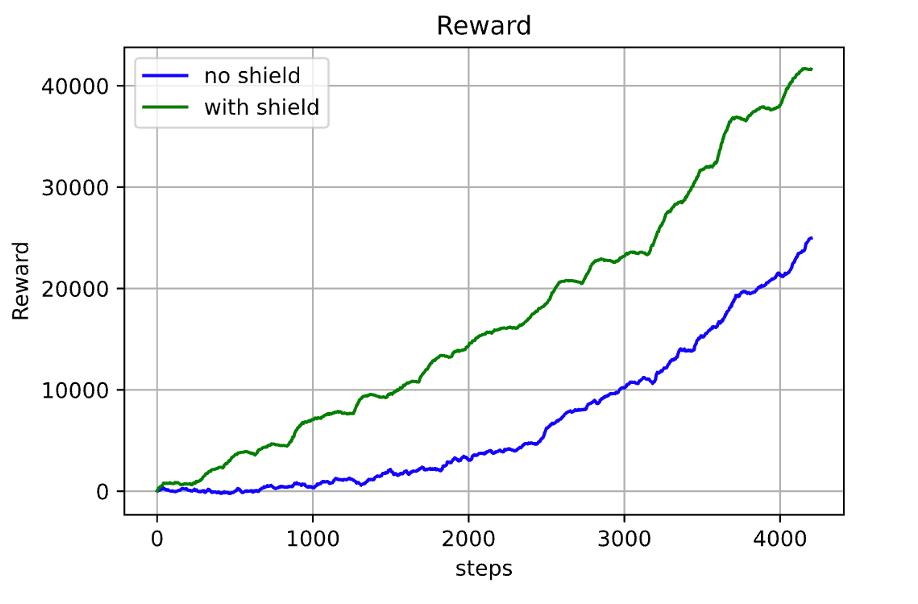}
\caption{Plot showing  the comparison of cumulative reward values with and without the safety shield, for the LTL specification $\varphi_2$.}
\label{fig:rewards}
\end{figure}

\vspace{-2mm}

\section{Demonstration} \label{sec:demonstration}

We now detail a UI we have developed for demonstration purposes. The UI is designed to be used by a network operations engineer who can specify safety intents, monitor tilts and their impact, and supervise the RL agent's operation. The initial screen of the UI depicts a geographic area with the available radio sites and cells. By selecting one of the cells, a new screen appears with the KPI values depicted on the left. On the right part of the page, one can see:
\begin{enumerate}
\item the MDP system model;
\item a list of available LTL intents;
\item BAs representing each of the intents;
\item the button ``Run safe RL" to run the simulation;
\item the switch ``with/without shield" for enabling the safety shield.
\end{enumerate}
The chosen actions on the MDP are depicted in blue, while the blocked actions by the shield are depicted in red. The user can view the training process and the optimal choice of actions that guarantee the satisfaction of given input as well as the block of unsafe actions. The current high level of detail in the UI is meant to illustrate the technology, and develop trust in the solution (which is crucial for AI systems which can sometimes appear inscrutable). It can be imagined that a production UI would instead show a summary of selected and blocked actions instead of large MDP models. The impact of the shield may also be viewed, and it is seen that the shield successfully blocks a proportion of unsafe states. \\

The simulation is executed on an urban environment with parameters as presented on Table \ref{sim_parameters}. The UEs are randomly positioned in the environment. Once the user positions and networks parameters are provided, the simulator computes the path loss in the urban environment using the Okomura-Hata propagation model \cite{theodore}. Examples of LTL tasks that can be given as input to the UI and the user can chose from the list, are given as follows:
\begin{enumerate}
\item $\varphi_1 = \square (\neg \text{sinrLow} \wedge \text{quaHigh} \wedge \text{covHigh})$, i.e., ``SINR, coverage and quality are never degradated together".
\item $\varphi_2 = (\Diamond \text{covHigh}) \wedge (\Diamond \text{quaHigh}) \wedge (\Diamond \neg \text{osHigh})$, i.e., ``antenna never overshoots and will eventually achieve high coverage and high quality".
\item $\varphi_3 = \square(\Diamond \text{covHigh}) \wedge (\Diamond \text{qualHigh})$, i.e., ``high coverage and high quality over all futures". The BA automaton of this formula is depicted in Fig. \ref{fig:ltl}. 
\end{enumerate}

\begin{figure*}[t]
\centering
\includegraphics[height = 8cm, width = 18cm]{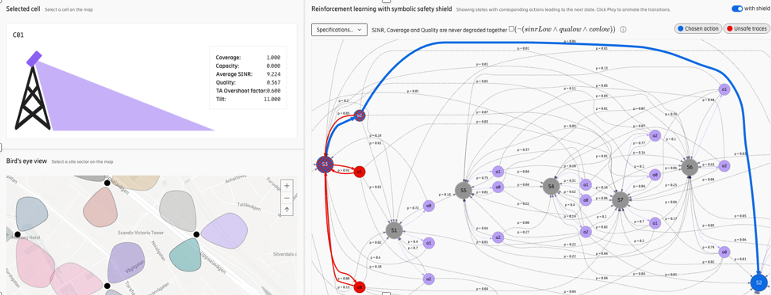}
\caption{The developed UI for the demonstration of the proposed approach. The user can specify an LTL formula, and can choose to view the resulting BA, and then inspect the evolution of the RL agent training, and the actions blocked by the safety shield.}\label{fig:UI}
\vspace{-0.2cm}
\end{figure*}

We now touch upon the efficacy of the safety shield. When we consider the specification $\varphi_1$, the number of safe states improves from $76\%$ safe states without the shield to $81.6\%$ safe states with the shield. In addition the cumulative reward value improves from 24,704 without the shield to 41,622.5 with the shield, an improvement of $68.5\%$. This may be attributed to the quality and coverage values remaining in desirable regions due to presence of the safety shield. 
We also show the reward plot for specification $\varphi_2$ in Fig. \ref{fig:rewards}. Notice that the cumulative reward is significantly improved with the shield as compared to without. This may be attributed to the coverage and quality KPI values being forced to better values by the safety shield. Further, there is an improvement in safety, with introduction of the shield leading to 639 unsafe states instead of 994 without the shield. These values can be further improved by making the shield more conservative. In this paper, the shield blocks an action if there is a $10\%$ probability its action leading to a violating trace. This may be reduced for example in the extreme case to any non-zero probability of leading to an unsafe trace. A demonstration video accompanying this paper can be found in:

\begin{table}[t!]
\vspace{3mm}
	\begin{center}
		\begin{tabular}{|c||c|}
			\hline
			Number of BSs  & $|\mathcal{R}| = 7$ \\
			\hline 
			Number of cells & $|\mathcal{C}| = 21$ \\		
			\hline
			Number of UEs & $|\mathcal{U}| = 2000$ \\
			\hline 
			Antenna height & $32 \ \rm meters$ \\
			\hline 
			Minimum dowtilt angle & $\theta_{\min} = 1^{\circ}$ \\
			\hline 
			Maximum downtilt angle & $\theta_{\max} = 16^{\circ}$ \\
			\hline 
			Discount factor & $\gamma = 0.9$ \\
			\hline
			Learning rate & $\eta = 0.1$ \\
			\hline 
			Batch size & $50$ \\
			\hline 
		\end{tabular}
	\end{center}
	\caption{The numerical values used for the training of the RL agent.}
	\label{sim_parameters}
\end{table}
\vspace{-2mm}
\begin{center}
https://youtu.be/cCDzaFd7D3k
\end{center}

\section{Conclusions} \label{sec:conclusions}

In this paper, we present an architecture for network KPIs optimization guided by user-defined specifications expressed in the rich LTL formalism. Our solution consists of MDP system dynamics abstraction, automata construction, cross product and model-checking techniques to block undesired actions that violate the specification with a given probability. Simulation results show that the approach is promising for network automation and control through deploying optimal strategies. In addition to this, a UI has been developed in order for a user to have interaction with all the steps of the proposed procedure for developing trust in the safe RL solution. 

Future research directions will be devoted towards applying the proposed framework in other telecom use cases as well as in robotics (motion planning). In addition, future work will focus on challenges in real-time model checking as the RL agent evolves, and on deriving guarantees for safety based on confidence measures for the MDP model.

\linespread{0.90}\selectfont
\bibliographystyle{ieeetr}
\bibliography{references}
\end{document}